\documentclass[final]{cvpr}

\usepackage{times}
\usepackage{epsfig}
\usepackage{graphicx}
\usepackage{amsmath}
\usepackage{amssymb}
\usepackage{url}
\usepackage{subfigure}
\usepackage{tablefootnote}
\usepackage{nopageno}
\usepackage{soul}

\usepackage[pagebackref=true,breaklinks=true,colorlinks,bookmarks=false]{hyperref}

\def\cvprPaperID{17} 
\def\confYear{CVPR 2021}

\begin{document}

\title{ChaLearn LAP Large Scale Signer Independent Isolated Sign Language Recognition Challenge: Design, Results and Future Research}


\author{Ozge Mercanoglu Sincan\\
Ankara University, Turkey\\
{\tt\small omercanoglu@ankara.edu.tr}
\and
Julio C. S. Jacques Junior\\
Computer Vision Center, Spain\\
{\tt\small jjacques@cvc.uab.cat}
\and \and
Sergio Escalera\\
University of Barcelona, Spain\\
Computer Vision Center, Spain\\
{\tt\small sescalera@ub.edu}
\and
Hacer Yalim Keles\\
Ankara University, Turkey\\
{\tt\small hkeles@ankara.edu.tr}
}

\maketitle

\begin{abstract}
The performances of Sign Language Recognition (SLR) systems have improved considerably  in recent years. However, several open challenges still need to be solved to allow SLR to be useful in practice. The research in the field is in its infancy in regards to the robustness of the models to a large diversity of signs and signers, and to fairness of the models to performers from different demographics. This work summarises the ChaLearn LAP Large Scale Signer Independent Isolated SLR Challenge, organised at CVPR 2021 with the goal of overcoming some of the aforementioned challenges. We analyse and discuss the challenge design, top winning solutions and suggestions for future research. The challenge attracted 132 participants in the RGB track and 59 in the RGB+Depth track, receiving more than 1.5K submissions in total. Participants were evaluated using a new large-scale multi-modal Turkish Sign Language (AUTSL) dataset, consisting of 226 sign labels and 36,302 isolated sign video samples performed by 43 different signers. Winning teams achieved more than 96\% recognition rate, and their approaches benefited from pose/hand/face estimation, transfer learning, external data, fusion/ensemble of modalities and different strategies to model spatio-temporal information. However, methods still fail to distinguish among very similar signs, in particular those sharing similar hand trajectories. 
\end{abstract}

\section{Introduction}

Sign/gesture recognition in the context of sign languages is a challenging research domain in computer vision, where signs are identified by simultaneous local and global articulations of multiple manual and non-manual sources, i.e. hand shape and orientation, hand motion, body posture, and facial expressions. Although the nature of the problems in this field is primarily similar to the action recognition domain, some peculiarities of sign languages make this domain specially challenging; for instance, for some pairs of signs, hand motion trajectories look very similar, yet local hand gestures look slightly different. On the other hand, for some pairs, hand gestures look almost the same, and signs are identified only by the differences in the non-manual features, i.e. facial expressions. In some cases, a very similar hand gesture can impose a different meaning depending on the number of repetitions. Another challenge is the variation of signs when performed by different signers, i.e. body and pose variations, duration variations etc. Also, variation in the illumination and background makes the problem harder, which is inherently problematic in computer vision. 

The performance of sign recognition algorithms have been improved considerably in recent years, mainly thanks to the release of associated datasets \cite{koller2020quantitative} and the development of new deep learning methodologies. Past works used to deal with data obtained in controlled lab environments, with a limited number of signers and signs. Recent works are dealing with more realistic and unconstrained settings and large scale datasets. In parallel, recent advancements in the domains of machine learning and computer vision, in particular deep learning, have pushed the state-of-the-art on the field substantially.
Still, several open challenges need to be solved to allow recognition systems to be useful in sign language, including signer independent evaluation, continuous sign recognition, fine-grain hand analyses, combination with face and body contextual cues, sign production, as well as model generalisation to different sign languages and demographics.

To motivate research in the field, we challenged researchers with a signer independent classification task using a novel large-scale, isolated Turkish Sign Language Dataset, named AUTSL ~\cite{Sincan:2020}. The video samples in AUTSL, containing variations in background and lighting, are performed by 43 signers. The challenge attracted a total of 191 participants, who made more than 1.5K submissions in total for the two challenge tracks. The RGB and RGB+Depth tracks (detailed in Sec.~\ref{sec:challengedesign}) received 1374 and 209 submissions, respectively, suggesting that the research community on SLR is currently paying more attention to RGB data, compared to RGB+Depth information. Moreover, top-winning solutions employed a wide variety of methods, such as the use of body/face/hand estimation/segmentation, different fusion/ensemble strategies and spatio-temporal modelling, external data and/or transfer learning, among others. 

The rest of this paper is organised as follows: in Sec.~\ref{sec:relatedwork}, we provide a short literature review. In Sec.~\ref{sec:challengedesign}, we present the challenge design, evaluation protocol, dataset and baseline. Challenge results and top-winning methods are discussed in Sec.~\ref{sec:challengeresults}. Finally, in Sec.~\ref{sec:conclusions}, we conclude the paper with a discussion and suggestions for future research.

\begin{table*}[htbp]
	\caption{Overview of isolated sign language/gesture datasets.}
	\centering
    \setlength\tabcolsep{3.4pt}
    \small
	\label{tab:datasets}
	\begin{tabular}{ |l|c|c|c|c|c|c| }
		\hline
		\textbf{Datasets} & \textbf{Year} &  \textbf{Signer independent} & \textbf{Modalities} & \textbf{\#Signs}  & \textbf{\#Signers}  & \textbf{\#Samples} \\ \hline
		RWTH  BOSTON50 \cite{zahedi2005combination}   & 2005  &  No &   RGB & 50 & 3 &  483 \\ \hline
		DGS \cite{cooper2012sign}   &  2012 &   No  &  RGB, depth & 40 & 15 & 3,000  \\ \hline
		GSL \cite{cooper2012sign}   & 2012  & No &  RGB & 20 & 6 &  840 \\ \hline
		Montalbano V1, V2 \cite{escalera2013chalearn, escalera2014chalearn}   & 2014  &  No &  RGB, depth, audio, user mask, skeleton & 20 & 27 & 13,858  \\ \hline
		DEVISIGN\cite{chai2015devisign}  &  2014 &  No &  RGB, depth & 2,000 & 8 &  24,000 \\ \hline
		PSL \cite{kapuscinski2015recognition}   &  2015 & No &  RGB, depth & 30 & 1 &  300 \\ \hline
	    LSA64\cite{ronchetti2016lsa64}  &  2016 & No &  RGB & 64 & 10 &  3,200 \\ \hline
		isoGD \cite{zahedi2005combination}   & 2016  &  Yes &  RGB, depth  & 249 & 21 &  47,933 \\ \hline
		MS-ASL \cite{joze2018ms}   & 2019  & Yes &  RGB & 1,000 & 222 & 25,513  \\ \hline
		CSL \cite{huang2018attention} & 2019  &  Yes &  RGB, depth, skeleton & 500 & 50 & 125,000 \\ \hline
		WLASL \cite{li2020word}       & 2020  & No &  RGB & 2,000 & 119 & 21,083   \\ \hline
		AUTSL \cite{Sincan:2020} & 2020  &  Yes & RGB, depth, user mask, skeleton  & 226 & 43 & 36,302  \\ \hline
	\end{tabular}
\end{table*}

\section{Related Work}~\label{sec:relatedwork}
Automatic sign language recognition has been an active area of research since early 90s. Early studies relied on using colored gloves or haptic sensors to segment and track hands \cite{fels1993glove, grobel1997isolated, mehdi2002sign}. However, intrusive methods that require wearing external gloves with some probes create practical difficulties in daily life and often limit the movements of the signers. Therefore, recent studies focus more on computer vision based solutions that use only cameras as the primary equipment for a solution. 

Early studies were trained and evaluated on small-scale datasets in terms of number of signs and signers, e.g., Purdue RVL-SLLL \cite{martinez2002purdue},  RWTH BOSTON50 \cite{zahedi2005combination}. In these studies, hand-crafted features, such as scale invariant feature transform (SIFT), histogram of oriented gradients (HOG) \cite{dardas2011real, han2009modelling}, were frequently used. After feature extraction, support vector machine (SVM) models or sequence models, such as Hidden Markov Models (HMMs) \cite{dardas2011real, zaki2011sign}, were used for classification. Similar to earlier works, some studies segmented hand regions before extracting the features, yet this time utilising computer vision based methods, like skin color detection, hand motion detection and trajectory estimation \cite{han2009modelling, yang2010chinese}. 

The emergence of Microsoft Kinect technology in 2010 enabled obtaining new data modalities, such as depth and skeleton, alongside RGB data sequence. New sets of small-scale multi-modal datasets (with less than 50 signs and 15 signers) were created using Kinect, such as DGS \cite{cooper2012sign}, GSL \cite{cooper2012sign} and PSL \cite{kapuscinski2015recognition}. In ChaLearn Looking at People (LAP) 2013 challenge, a multi-modal Italian gesture dataset, Montalbano V1, was released~\cite{escalera2013chalearn}, including RGB, depth, user mask, skeletal model, and audio. It contains 20 gestures and approximately 14,000 samples performed by 27 different signers in total. In ChaLearn LAP 2014 challenge, an enhanced version of the dataset, i.e. Montalbano V2, was released~\cite{escalera2014chalearn}. Although there is only 20 different signs, Montalbano gesture dataset contains more samples and more variance in the video recordings than previously released datasets. In 2014, a large scale isolated Chinese Sign Language that is named as DEVISIGN was released \cite{chai2015devisign}. It consists of 2,000 signs that are performed by 8 signers. The videos were recorded in a lab environment with a controlled background. With the emergence of multiple modalities, researchers worked on different fusing techniques using the features extracted from these modalities, e.g., early, intermediate or late fusion, to get more robust results \cite{wan2013one, neverova2014multi, pigou2014sign, wu2016deep}. Moreover, recent advances prompted researchers to extract features using deep learning based models, instead of using hand-crafted features. Some works preferred using both manually extracted features and deep learning based features together \cite{neverova2014multi, wu2016deep}.

In 2016, Chalearn LAP RGB-D Isolated Gesture Recognition (IsoGD) dataset was released \cite{wan2016chalearn}. It was planned to challenge researchers for high performance automatic classification in ``large-scale'' and ``signer independent'' evaluation settings, which means that the samples in the test set are performed by different signers from the train set. In this dataset, there are 249 gestures that are performed by 21 different signers; each class contains approximately 200 RGB and depth videos. 
In the related years, commonly, 2D-CNN based models were used for feature extraction and sequence models, such as RNN, LSTM, GRU, HMM, were used for encoding temporal information \cite{sincan2019isolated, koller2019weakly, rastgoo2020video, tur2021evaluation}. Recent developments in action recognition have also contributed significantly to the recognition of signs in sign languages. Using and fine-tuning 3D-CNN models, e.g., C3D \cite{tran2015learning}, I3D \cite{8099985}, pre-trained on large action recognition datasets helped achieving higher accuracy rates compared to 2D-CNNs 
\cite{li2016large,joze2018ms, huang2018attention, adaloglou2020comprehensive}. 

In recent years, a number of large-scale isolated sign language datasets have been released, with large vocabulary sizes, large number of samples performed by many signers, e.g., MS-ASL \cite{joze2018ms}, CSL \cite{huang2018attention} and WLASL~\cite{li2020word}. MS-ASL provided 1,000 signs with 222 signers in signer independent setting. It was collected from a public video sharing platform. CSL is a multi-modal Chinese Sign Language dataset that consists of 500 signs performed by 50 different signers, arranged for signer independent evaluations. It contains RGB, depth, and skeleton data modalities. WLASL consists of 2,000 signs performed by 119 signers. It was collected from sign language websites. Although each of these datasets has several different challenges, video samples usually have plain backgrounds and data is collected in a controlled setting. Table \ref{tab:datasets} provides an overview of the available isolated sign language/gesture datasets.

In the context of this challenge, a new large-scale, multi-modal Turkish Sign Language dataset, AUTSL \cite{Sincan:2020}, is utilized in a signer independent evaluation setting. Different from the other large-scale datasets, it contains a variety of 20 different backgrounds obtained from indoor and outdoor environments, with several challenges (detailed in Sec.~\ref{sec:dataset}).

\section{Challenge Design}\label{sec:challengedesign}
The challenge\footnote{Challenge Webpage: \scriptsize\url{http://chalearnlap.cvc.uab.es/challenge/43/description/}} focused on isolated Sign Language Recognition (SLR) from signer independent non-controlled RGB+D (depth) data, involving a large number of sign categories ($>$200, detailed in Sec.~\ref{sec:dataset}). It was divided into two different competition tracks, i.e., RGB\footnote{\scriptsize\url{https://competitions.codalab.org/competitions/27901}} and multimodal RGB+D\footnote{\scriptsize\url{https://competitions.codalab.org/competitions/27902}}. The only restriction was that depth data was not allowed in any format and stage of training in RGB track. The participants were free to join any of these tracks. Both modalities have been temporally and spatially aligned. Each track was composed of two phases, i.e., development and test phase. At the development phase, public train data was released and participants submitted their predictions with respect to a validation set. At the test (final) phase, participants were requested to submit their results with respect to the test data. Participants were ranked, at the end of the challenge, using the test data.

The challenge ran from 22 December 2020 to 11 March 2021 through Codalab\footnote{\scriptsize\url{https://codalab.org/}}, a powerful open source framework for running competitions that involve result or code submission. It attracted a total of 191 registered participants, 132 in RGB track and 59 in RGB+D track. During development phase we received 1317 submissions from 39 teams in the RGB track, and 176 submissions from 15 teams in the RGB+D track. At the test (final) phase, we received 57 submissions from 23 teams in the RGB track, and 33 submissions from 14 teams in the RGB+D track. The reduction in the number of submissions from the development to the test phase is explained by the fact that the maximum number of submissions per participant on the final phase was limited to 3, to minimise the change of participants to improve their results by try and error.

It is important to note that the challenge was designed to deal with the submission of results (and not code). Participants submitted only their prediction files containing one label for each video. Therefore, participants were required to share their codes after the end of the challenge so that the organisers could validate their results in a ``code verification stage''. At the end of the challenge, top ranked methods (discussed in Sec.~\ref{sec:winningmethods}) passing the code verification stage (e.g., they publicly released their codes and the organisers were able to reproduce the results) were announced as top winning solutions.

\subsection{The Dataset}\label{sec:dataset}

\begin{figure*}[htbp] 
	\centering    
    \includegraphics[width=0.9\textwidth]{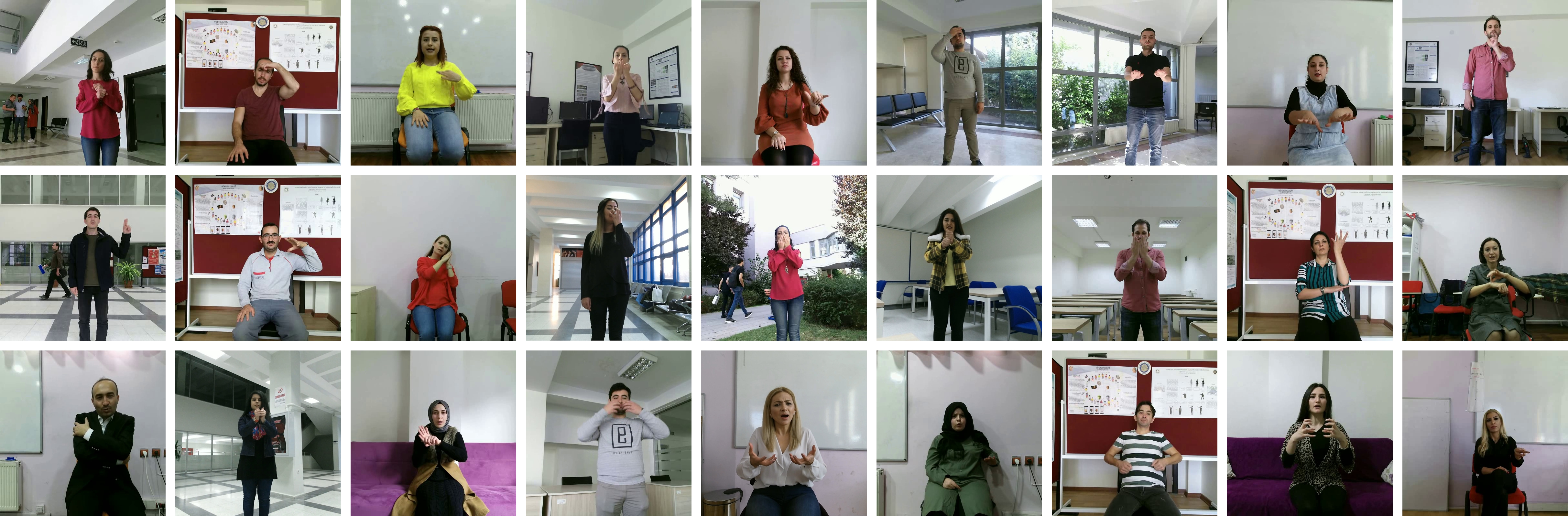}
	\caption{Some screenshots from the AUTSL~\cite{Sincan:2020} dataset.}
	\label{fig:dataset}
\end{figure*}

AUTSL ~\cite{Sincan:2020} is a large-scale, signer independent, multi-modal dataset that contains isolated Turkish sign videos. It contains 226 signs that are performed by 43 different signers.  The dataset is recorded with Microsoft Kinect V2 and contains RGB, depth, user mask, and skeleton data. Only RGB and depth modalities are released within the scope of the challenge. Some clipping and resizing operations are applied to RGB and depth data and video frames are resized to 512 x 512 pixel resolution. The average number of frames per video is $60\pm10$, while the frame rate for video is 30 frame per second (fps).

In the associated publication \cite{Sincan:2020}, while AUTSL test split was designed as a signer independent set; validation split was generated using a random split, i.e. 15\% of the train data. For this challenge, we  split the training set to create a signer independent validation set, making all sets signer independent. We selected 31 signers for training, 6 signers for validation, and 6 signers for testing. In this setting, training set contains 28,142, validation set contains 4,418, and test set contains 3,742 video samples.   
 AUTSL is a balanced dataset according to the sign distribution, i.e. each sign contains approximately the same number of samples ($\sim$160). The train, validation and test set contain approximately 124, 19, and 17 samples per sign, respectively. Signs are selected from the daily spoken vocabulary. They cover a wide variety in terms of hand shape and hand movements; some signs are performed only with one hand while some with both hands, in some signs hands occlude each other or parts of the face.
 We depict examples of different backgrounds and signers from the dataset in Fig.~\ref{fig:dataset}.

\textit{\textbf{Challenges:}} The dataset has various challenges, including lighting variability, different postures of signers, dynamic backgrounds, such as moving trees, or moving people behind the signer, high intra-class variability and inter-class similarities. 
In order to provide a basis for signer independent recognition systems, train, val, and test splits include different signers. The dataset contains 20 different backgrounds with several challenges. The test set contains 8 different backgrounds, 3 of which are not included in the training or validation sets. Another challenge is the inter-class similarity of signs; some signs contain exactly the same hand gesture, but differing only by the number of repetitions of the same gesture. Also, some signs are quite similar in terms of hand shape, hand orientation, hand position or hand movement; there is only subtle differences. 

\textit{\textbf{Limitations:}} The fact that the society is right-handed in general is also reflected in the distribution in AUTSL. Only 2 of the signers are left-handed out of 43 signers. Therefore, there is a bias towards the right handed signers in the dataset. Furthermore, female signers are more dominant, almost 3:1 ratio, in the dataset; 10 of the signers are men and 33 are women. The ages of our signers range from 19 to 50, and the average age of all signers is 31. In other words, there are no child or elderly signers. Another point that can be considered a source of bias in the dataset is the distribution of skin color, as there is no signer with dark skin. Although these limitations exist, we believe the challenge we have organised can help to advance the state-of-the-art on the field, as well as to promote either the design of new dataset or the development of novel methodologies that can deal with the aforementioned limitations.

\subsection{Evaluation Protocol}
To evaluate the performances of the models, we use the recognition rate, $r$, as defined in previous ChaLearn LAP challenges \cite{wan2016chalearn}.

\begin{equation}
\label{equ:r}
r=\frac{1}{n} \sum_{i=1}^{n}f(p_{i}, y_{i}),
\end{equation}
where $n$ is the total number of samples; $p_{i}$ is the predicted label for the $i^{th}$ sample; $y_{i}$ is the true label for the $i^{th}$ sample; $f(.)$ is $1$ when $p_{i}= y_{i}$, $0$ otherwise.

\subsection{The Baseline}

In order to set a baseline, several deep learning based models are trained and evaluated on AUTSL dataset. In the baseline method \cite{Sincan:2020}, 2D-CNNs are used to extract spatial features. Then, a Feature Pooling Module (FPM) \cite{sincan2019isolated} is placed on top of the last CNN layer. The idea behind FPM layers is to increase the field-of-views by using different dilated convolutions. In order to capture temporal information bidirectional LSTM (BLSTM) is used. A temporal attention mechanism is integrated to BLSTM in order to select the most effective video frames in classification. 

The methods used in RGB and RGB+D track are basically the same, with minor modifications. Since the depth data is represented as a single channel gray-scale image for each frame, the same depth data is repeated into three color channels. Then, RGB and depth modalities are given as inputs to the two parallel CNN models that share the same parameters. After generating two feature matrices, i.e. one for the RGB data and one for the depth data, these feature matrices are concatenated at the end of the FPM layer.

In contrast to some top-winning solutions (detailed in Sec.~\ref{sec:challengeresults}), our baseline did not consider any face/hand/body detection or segmentation technique, nor additional modalities such as pose keypoints, optical flow, nor external data. 


\section{Challenge Results and Winning Methods}~\label{sec:challengeresults}

\begin{figure*}[htbp]
	\centering
	\begin{subfigure}[RGB Track]
	{\includegraphics[width=0.48\linewidth]{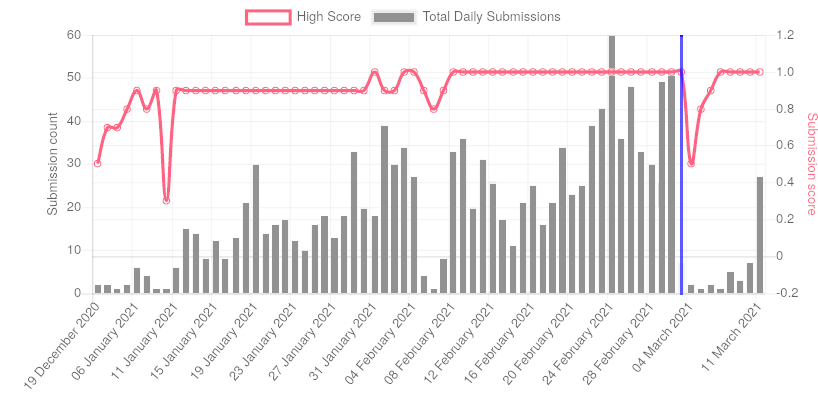}}
	\end{subfigure}
	\begin{subfigure}[RGB+D Track]
	{\includegraphics[width=0.48\linewidth]{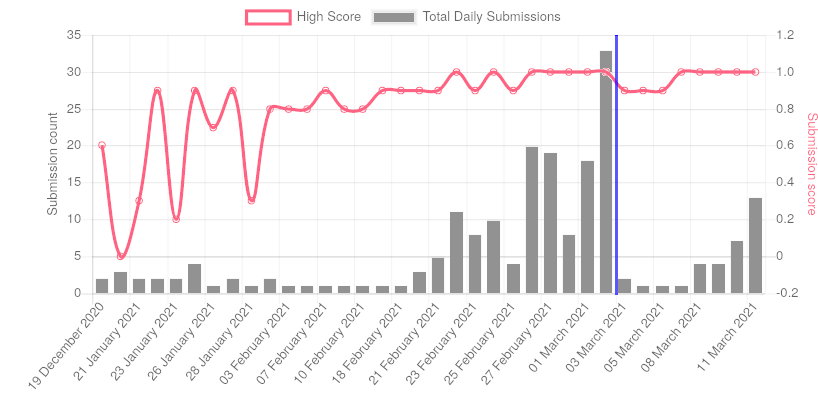}}
	\end{subfigure}
\caption{Challenge evolution with respect to the number of submissions and best obtained score, per day and per track. The blue line indicates the end of the development phase and the start of the test phase.}	
	\label{res:challengeevolution}
\end{figure*}

\subsection{The Leaderboard}\label{sec:leaderboard}
Results obtained by the top-10 winning solutions (in addition to the \textit{Baseline}) at the test phase, for the RGB and RGB+D tracks, are reported in Table~\ref{res:top-10:RGB}. The main observation from the table is that RGB+D and RGB results do not significantly differ, suggesting that state-of-the-art methods are obtaining highly accurate results without the use of depth information, at least on the adopted AUTSL~\cite{Sincan:2020} dataset. Later in Sec.~\ref{sec:analysisofresults}, we analyse possible causes why some samples were not properly recognised by the top winning solutions, which could be used to guide future research directions on the field.

\begin{table}[htbp]
\centering
\setlength\tabcolsep{2.0pt}
\small
\caption{Codalab leaderboards of RGB and RGB+D Tracks. Methods that passed the code verification stage are highlighted in bold.}
\begin{tabular}{|c|l|c|l|c|l|c|}
\cline{1-3} \cline{5-7}
\multicolumn{3}{|c|}{\textbf{RGB Track}} & \textbf{} & \multicolumn{3}{c|}{\textbf{RGB+D Track}} \\ \cline{1-3} \cline{5-7} 
Rank & \multicolumn{1}{c|}{Participant} & Rec. Rate &  & Rank & \multicolumn{1}{c|}{Participant} & Rec. Rate \\ \cline{1-3} \cline{5-7} 
1 & \textbf{smilelab2021} & 0.9842 &  & 1 & \textbf{smilelab2021} & 0.9853 \\ \cline{1-3} \cline{5-7} 
2 & wz & 0.9834 &  & 2 & wz & 0.9834 \\ \cline{1-3} \cline{5-7} 
3 & \textbf{rhythmblue} & 0.9762 &  & 3 & \textbf{rhythmblue} & 0.9765 \\ \cline{1-3} \cline{5-7} 
4 & \_Bo\_ & 0.9743 &  & 4 & \textbf{wenbinwuee} & 0.9669 \\ \cline{1-3} \cline{5-7} 
5 & \textbf{wenbinwuee} & 0.9655 &  & 5 & lin.honghui & 0.9567 \\ \cline{1-3} \cline{5-7} 
6 & deneme4 & 0.9626 &  & 6 & ly59782 & 0.9548 \\ \cline{1-3} \cline{5-7} 
7 & \textbf{jalba} & 0.9615 &  & 7 & Bugmaker & 0.9396 \\ \cline{1-3} \cline{5-7} 
8 & xz & 0.9596 &  & 8 & \textbf{m-decoster} & 0.9332 \\ \cline{1-3} \cline{5-7} 
9 & wuyongfa & 0.9580 &  & 9 & papastrat & 0.9172 \\ \cline{1-3} \cline{5-7} 
10 & adama & 0.9578 &  & 10 & xduyzy & 0.9086 \\ \cline{1-3} \cline{5-7} 
23 & \textit{Baseline} & 0.4923 &  & 14 & \textit{Baseline} & 0.6203 \\ \cline{1-3} \cline{5-7} \end{tabular}
\label{res:top-10:RGB}
\end{table}

Fig.~\ref{res:challengeevolution} illustrates the evolution of the challenge with respect to the number of submissions and highest score obtained for each day and competition track. Different observations can be made from these plots: 1) the participants were much more active in the RGB track, also reinforced by the number of registered participant on this track, suggesting that the research community is paying more attention on SLR from RGB data if compared to RGB+D information; 2) the number of submissions increases close to the end of each phase (development phase finished on 3rd of March and the test phase finished on 11th of March), suggesting that participants were struggling to improve their results to obtain a better rank position. 

\subsection{Top Winning Approaches}\label{sec:winningmethods}

\begin{table*}[htbp]
\centering
\small
\caption{General information about the top-3 winning approaches.}
\begin{tabular}{|l|c|c|c|c|c|c|}
\hline
\multicolumn{1}{|c|}{\textit{Participant}} & \multicolumn{2}{c|}{\textbf{\begin{tabular}[c]{@{}c@{}}top-1:\\ smilelab2021\end{tabular}}} & \multicolumn{2}{c|}{\textbf{\begin{tabular}[c]{@{}c@{}}top-2:\\ rhythmblue\end{tabular}}} & \multicolumn{2}{c|}{\textbf{\begin{tabular}[c]{@{}c@{}}top-3:\\ wenbinwuee\end{tabular}}} \\ \hline
\multicolumn{1}{|c|}{\textit{Feature/Track}} & \multicolumn{1}{l|}{\textit{RGB}} & \multicolumn{1}{l|}{\textit{RGB+D}} & \multicolumn{1}{l|}{\textit{RGB}} & \multicolumn{1}{l|}{\textit{RGB+D}} & \multicolumn{1}{l|}{\textit{RGB}} & \multicolumn{1}{l|}{\textit{RGB+D}} \\ \hline
Depth information either during training or testing stage & - & $\surd$ & - & $\surd$ & - & $\surd$ \\ \hline
Pre-trained models & $\surd$ & $\surd$ & $\surd$ & $\surd$ & $\surd$ & $\surd$ \\ \hline
External data & $\surd$ & $\surd$ & - & - & - & - \\ \hline
Regularization strategies/terms & $\surd$ & $\surd$ & - & - & - & - \\ \hline
Handcrafted features & - & $\surd$ & - & - & - & - \\ \hline
Face/hand/body detection, alignment or segmentation & - & - & $\surd$ & $\surd$ & $\surd$ & $\surd$ \\ \hline
Pose estimation & $\surd$ & $\surd$ & $\surd$ & $\surd$ & - & - \\ \hline
Fusion of modalities & $\surd$ & $\surd$ & $\surd$ & $\surd$ & - & - \\ \hline
Ensemble models & $\surd$ & $\surd$ & $\surd$ & $\surd$ & $\surd$ & $\surd$ \\ \hline
Spatio-temporal feature extraction & $\surd$ & $\surd$ & $\surd$ & $\surd$ & $\surd$ & $\surd$ \\ \hline
Explicitly classify any attribute (e.g. gender) & - & - & - & - & - & - \\ \hline
Bias mitigation technique (e.g. rebalancing training data) & - & - & - & - & - & - \\ \hline
\end{tabular}
\label{tab:featurestop3}
\end{table*}

This section briefly presents the top winning approaches of both tracks. More concretely, the top-3 methods that passed the code verification stage (see Table~\ref{res:top-10:RGB}). Table~\ref{tab:featurestop3} shows some general information about the top-3 winning approaches. 
As it can be seen from Table~\ref{tab:featurestop3}, common strategies employed by top-winning solutions are transfer learning, external data, face/hand detection and pose estimation, fusion of modalities and ensemble models as well as different strategies to model spatio-temporal information.

\subsubsection{Top-1: smilelab2021}
Inspired by the recent development of whole-body pose estimation~\cite{Sheng:eccv:2020}, the \textit{smilelab2021}\footnote{Code: \scriptsize\url{https://github.com/jackyjsy/CVPR21Chal-SLR}} team proposed to recognise sign language based on the whole-body key points and features. The recognition results are further ensembled with other modalities of RGB and optical flows to further improve the accuracy. The top-1 winning solution~\cite{jiang2021skeleton} proposed to use whole-body pose keypoints to recognise sign language via a multi-stream Graph Convolutional Network (GCN) model. 

A total of 133-points including face, hand, body and foot are extracted from the input images. They are used in their GCN network as skeleton modality. Features extracted from pretrained whole-body pose estimation are used as another modality. The keypoints are also used to crop frames in
other modalities (RGB and optical flow). As base model, Resnet2+1d pretrained on Kinetics~\cite{carreira2018short} dataset is used. For RGB modality, they pre-trained their models on Chinese Sign Language dataset~\cite{7552950} before training on the challenge dataset. Label smoothing and weight decay were used as regularization during training on both tracks. In the RGB+D track, they extracted HHA~\cite{jiang2021skeleton} features from depth video as another modality, referred to as handcraft features. HHA features encode depth information and are generated using a RGB-like 3-channel output, where HHA stand for ``Horizontal disparity'', ``Height above the ground'', and ``Angle normal makes with''. Since multiple modalities are considered (skeleton keypoints, skeleton features, RGB and optical flow - and HHA and depth flow in the case of RGB+D track), they adopted a late fusion technique where the output of the last fully-connected layers is kept, before softmax, associating weights to them and sum them up with weights as a final predicted score. Those weights serve as hyper-parameters and are tuned based on the accuracy on validation set. 


\subsubsection{Top-2: rhythmblue}

The \textit{rhythmblue}\footnote{Code: \scriptsize\url{https://github.com/ustc-slr/ChaLearn-2021-ISLR-Challenge}} team proposed an ensemble framework composed of multiple neural networks (e.g., I3D, SGN) to conduct isolated sign language recognition, also taking into account pose, hand and face patch-based information. The networks are trained separately for different cues. For patch sequence of full-frame, hands and face, 3D-CNNs are used to model the spatio-temporal information. For pose data, GCN-based method is selected to capture the skeleton correlation. During ensemble stage, late fusion is adopted for final prediction.

More concretely, an upper-body patch is obtained according to a bounding box estimation of the signer, given by MMDetection~\cite{mmdetection} and the joint positions estimated by HRNet~\cite{Sun_2019_CVPR}. Full body joint positions are extracted with MMPose~\cite{mmpose2020}. Hand and face regions are obtained from keypoint positions. Five types of data are generated, i.e., full-body patch, left-hand patch, right-hand patch, face patch and full-body pose. To process full-body patch, left-hand patch and right-hand patch, separately I3D~\cite{8099985} networks are used. A SlowFast~\cite{Feichtenhofer_2019_ICCV} Network is used for full-body patch. To process full-body pose, SGN~\cite{Zhang_2020_CVPR} is used. During
inference, all outputs are summed before the SoftMax layers of the above networks with weights. Then, the category with the largest activation is selected. In the case of RGB+D track, I3D-Depth is used, with a pretrained model provided by I3D-Kinectics-Flow~\cite{8099985}. 


\subsubsection{Top-3: wenbinwuee}
The \textit{wenbinwuee}\footnote{Code: \scriptsize\url{https://github.com/Koooko96/Chalearn2021code}} team used RGB, Depth information (in RGB+D track), optical flow and human segmentation data to train several models using SlowFast~\cite{Feichtenhofer_2019_ICCV}, SlowOnly~\cite{Feichtenhofer_2019_ICCV} and TSM~\cite{9008827}. Results are late fused to get a final prediction. For the RGB+D track, results are obtained by fusing the RGB+D models prediction scores with the RGB track results. 


\begin{figure*}[htbp] 
	\centering    
    \includegraphics[width=1\textwidth]{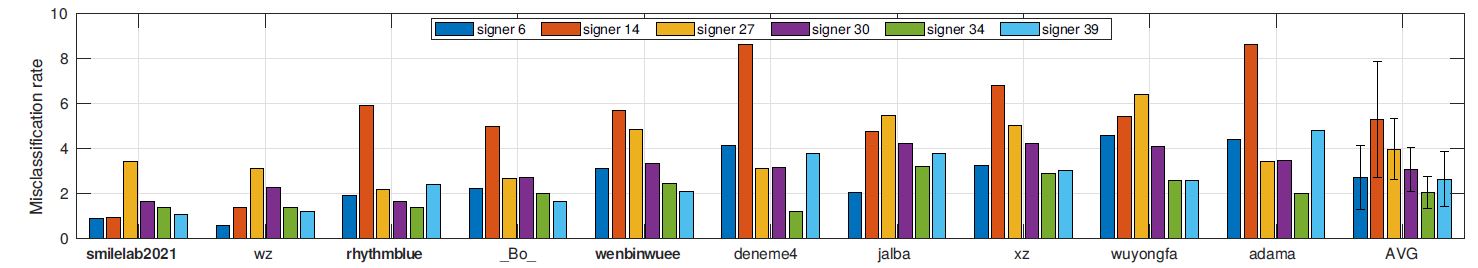}
	\caption{Misclassification rate (\%) per signer on the test set, given the top-10 teams (shown in Table~\ref{res:top-10:RGB}) in the RGB track. Average misclassification rate is also reported. In bold, the top-3 winning solutions that passed the code verification stage.}
	\label{fig:wrong_percentage_signers}
\end{figure*}
\subsection{What challenge the models the most?}\label{sec:analysisofresults}

In this section, we analyse the prediction files submitted at the test phase on both competition tracks and for the top-10 teams shown in Table~\ref{res:top-10:RGB}, and discuss some particularities that challenge the methods the most. For instance, we show the samples that were more frequently wrongly classified given the signer or sign IDs, which could indicate a weakness of the evaluated methods or any issue in the adopted dataset (e.g., high inter-class similarity), that could suggest future research.

Fig.~\ref{fig:wrong_percentage_signers} shows misclassification rates in the RGB track for each signer in the test set. No significant differences were observed for the results that are obtained for RDB+D track. The first 10 segments show the results of the top-10 teams shown in Table~\ref{res:top-10:RGB}, and the last segment shows the average misclassification rates per signer. As it can be seen, different teams misclassified different signers without a clear pattern, suggesting that there was not a particular signer (or set of signers) that could be considered an outlier. Nevertheless, if we take the average misclassification as reference, we can observe that Signer 14 and 27 were the ones more frequently misclasified (which was not the case of the top-1 winning solution, at least for Signer 14).
One possible explanation for these cases is the high intra-class variability, which imposes an additional challenge for generalisation.

\begin{figure}[htbp] 
	\centering
	\begin{subfigure}[RGB Track]
    {\includegraphics[width=0.95\columnwidth]{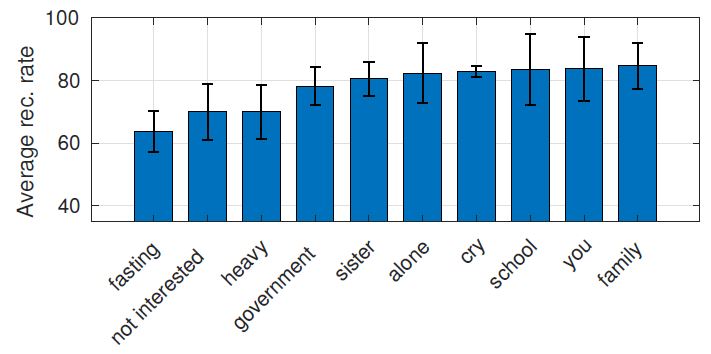}}
	\end{subfigure}
	\begin{subfigure}[RGB+D Track]
    {\includegraphics[width=0.95\columnwidth]{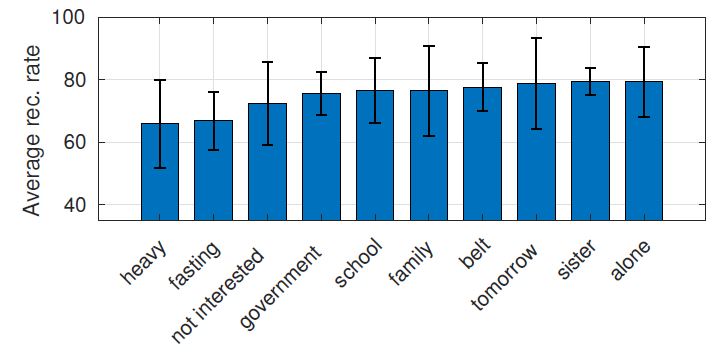}}
	\end{subfigure}	
\caption{Average recognition rate (\%) of top-10 misclassified signs (test set) given the top-10 teams (shown in Table~\ref{res:top-10:RGB}).}
	\label{fig:worst5sign}
\end{figure}

In Fig.~\ref{fig:worst5sign}, we show the average recognition rate of top-10 misclassified signs on the test phase, given the submitted files of top-10 teams (shown in Table~\ref{res:top-10:RGB}). As it can be seen, the set of signs with lower recognition rates (e.g., top-5) are more or less the same in both tracks, suggesting that RGB and RGB+depth are providing similar information to solve the task on those cases, or maybe that the complementarity of RGB and depth are not being fully exploited.

\begin{figure}[htbp] 
	\centering
	\begin{subfigure}[\{heavy \textit{vs.} lightweight\}]
    {\includegraphics[width=0.9\columnwidth]{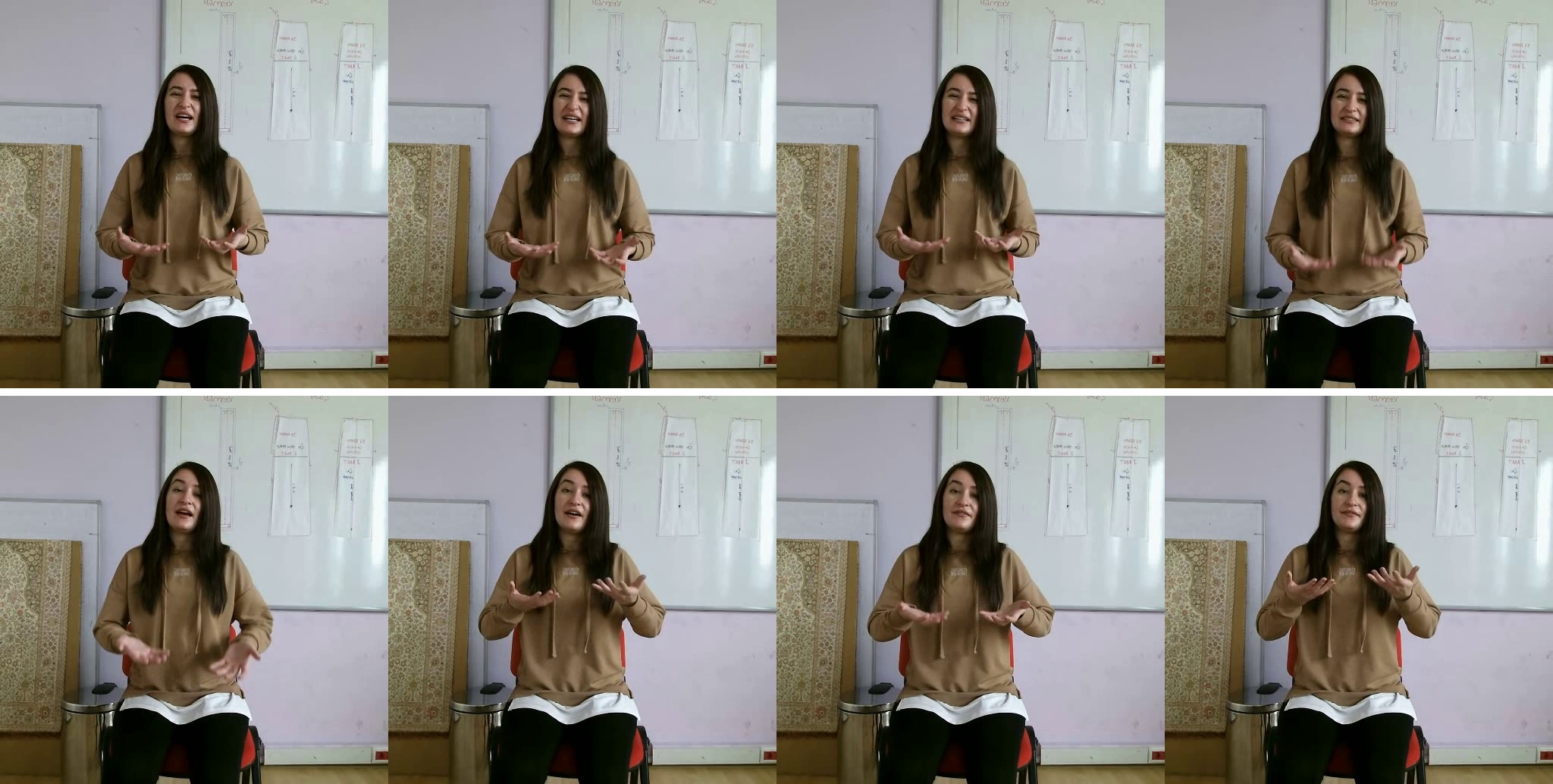}}
	\end{subfigure}
	\begin{subfigure}[\{not interested \textit{vs.} why\}]
    {\includegraphics[width=0.9\columnwidth]{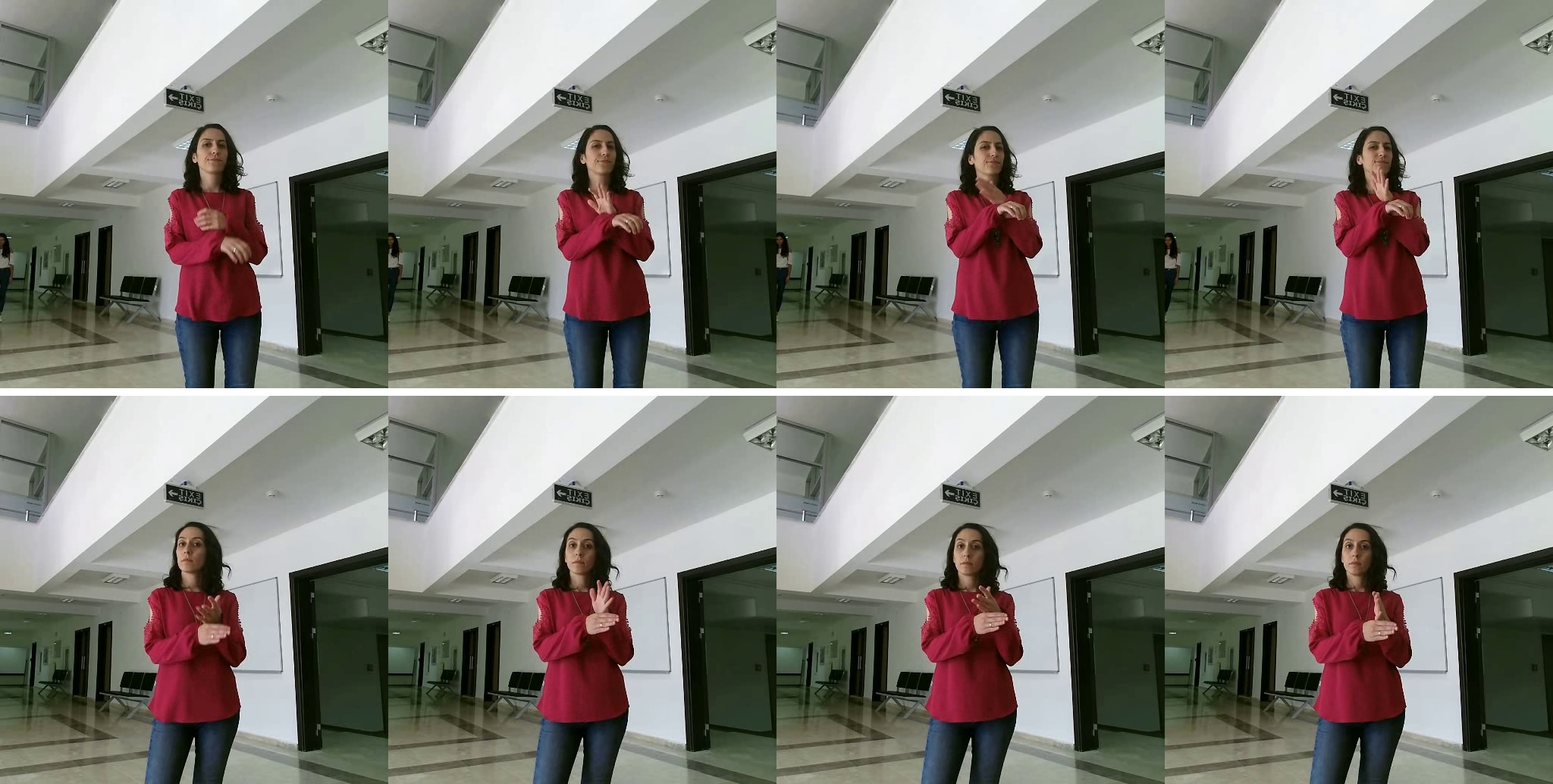}}
	\end{subfigure}
    \begin{subfigure}[\{fasting \textit{vs.} school \textit{vs.} soup\}]{\includegraphics[width=0.9\columnwidth]{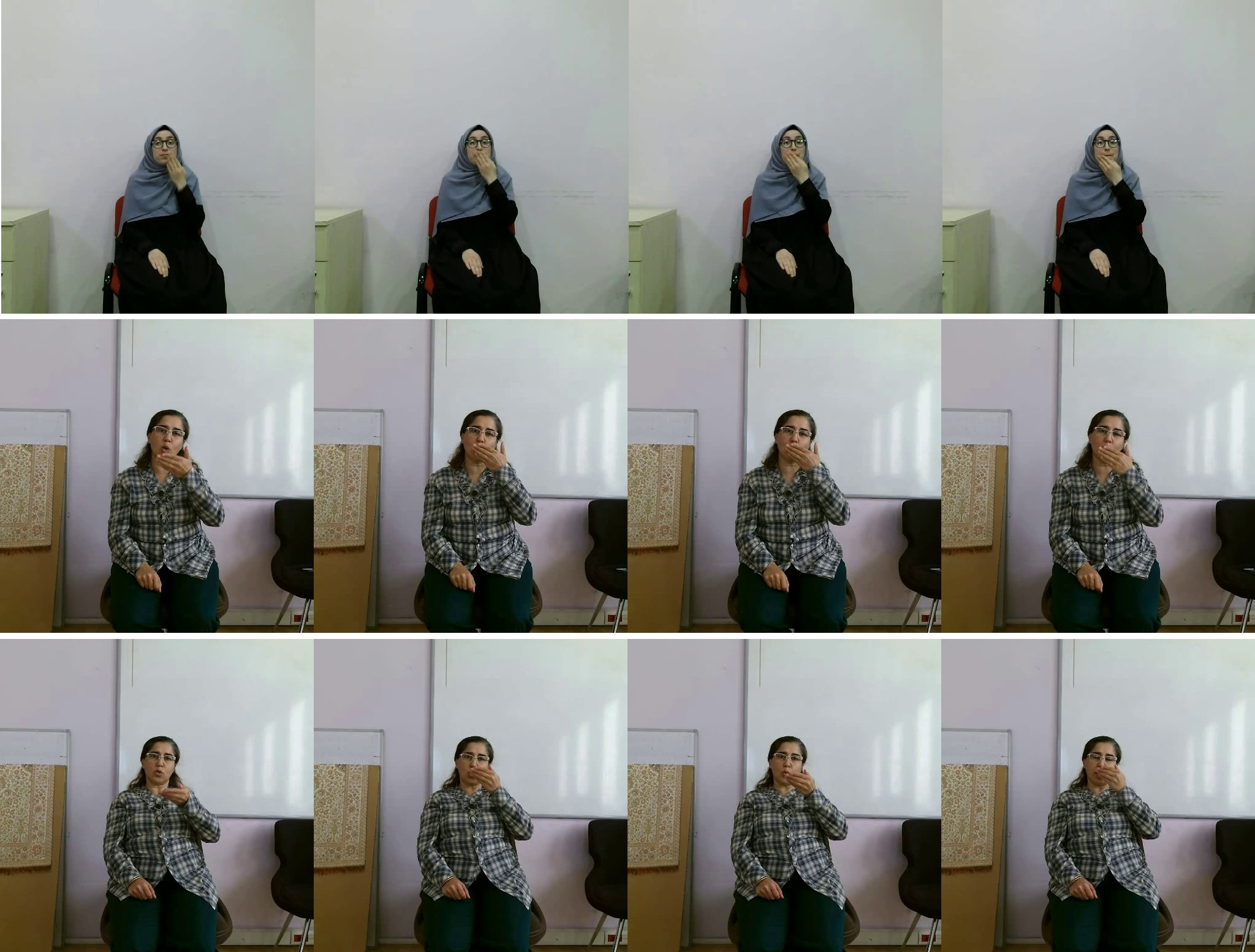}}
	\end{subfigure}
	\begin{subfigure}[\{government \textit{vs.} Ataturk\}]
    {\includegraphics[width=0.9\columnwidth]{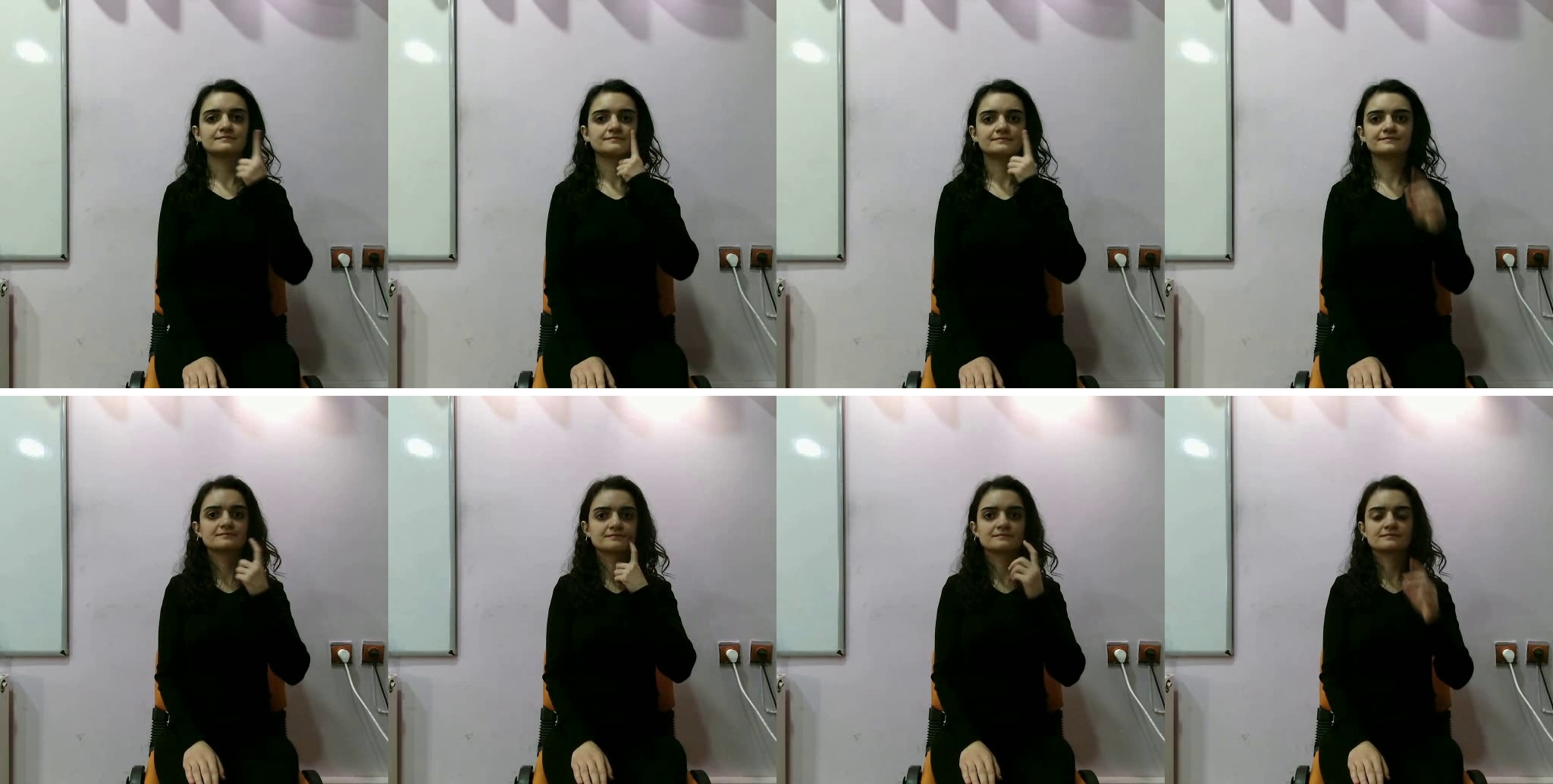}}
	\end{subfigure}	
\caption{The most frequently confused sign pairs in both RGB and RGB+D track. Each row displays a sign video sample summarised in 4 frames.}
	\label{fig:similarSigns}
\end{figure}

By analysing a confusion matrix of the signs and submitted predictions of both tracks, we observed that the most frequently confused sign pairs by participants methods in the AUTSL~\cite{Sincan:2020} dataset are: \{heavy \textit{vs.} lightweight\}, \{fasting \textit{vs.} school\}, \{not interested \textit{vs.} why\}, \{school \textit{vs.} soup\} and \{government \textit{vs.} Ataturk\}. The reason behind such misclassifications are due to the high similarity of local and global hand gestures in these signs, illustrated in Fig.~\ref{fig:similarSigns}. In some signs, there is only a subtle difference in the position of the hand, e.g., government and Ataturk. While in the sign of government the index finger touches under the eye, in the sign of Ataturk it touches the cheek. In some signs, there is only a subtle difference in movement of hands, e.g., school and soup. In the sign of soup the hand moves a little more from the bottom up. In some signs, facial expression also contains an important clue for the meaning of the sign, e.g., heavy.

\section{Conclusions}\label{sec:conclusions}

This work summarised the ChaLearn LAP Large Scale Signer Independent Isolated SLR Challenge. The challenge attracted more then 190 participants in two computational tracks (RGB and RGB+D), who made more than 1.5K submissions in total. We analysed and discussed the challenge design, top winning solutions and results. Interestingly, the challenge activity and results on the different tracks showed that the research community in this field is currently paying more attention to RGB information, compared to RGB+Depth data. Top winning methods combined hand/face detection, pose estimation, transfer learning, external data, fusion/ensemble of modalities and different strategies to model spatio-temporal information. 

We believe that future research directions should move at least in two different lines, that is, on the development of novel large-scale and public datasets, and on the research and development of methods that are both fair and accurate. Fairness is an emergent topic in computer vision and machine learning, and new datasets should include people from different ages, gender, skin tones, demographics, among others, with the goal of having as much as possible balanced distributions given the different attributes. Moreover, continuous sign language seems to be a logical next stage in order to do research on begin-end of sign detection and to include of a higher level of language semantics in the recognition process. The inclusion and analysis of context and spatio-temporal attention mechanisms could be helpful for discriminating very similar signs.
On the other hand, models are benefiting from state-of-the-art approaches developed for other purposes to achieve state-of-the-art performance. The fusion of different modalities and models seems to be a key to advance the research on this field. It should be noticed that the top-2 winning solutions benefited from Graph Convolutional Networks (GCN), which demonstrated to be very useful to model spatio-temporal information. Furthermore, up to date self-attention strategies have not been fully exploited in sign language, and its usage could benefit the spatio-temporal learning of signs.

Finally, future work should also consider paying more attention to explainability/interpretability, so that the results obtained by different models could be easily explained and interpreted. This is key to understand what part or components of the model are more relevant to solve a particular problem, or to explain possible sources of bias or misclassification.  




\section*{Acknowledgments}
This work has been partially supported by the Scientific and Technological Research Council of Turkey (TUBITAK) project 217E022, the Spanish project PID2019-105093GB-I00 (MINECO/FEDER, UE) and CERCA Programme/Generalitat de Catalunya, and ICREA under the ICREA Academia programme.

{\small
\bibliographystyle{ieee_fullname}
\bibliography{camera_ready}
}

\end{document}